\documentclass[10pt,twocolumn,letterpaper]{article}

\usepackage[pagenumbers]{cvpr}

\usepackage{cuted}%

\usepackage{currfile}%

\usepackage{caption}%

\makeatletter
\renewcommand{\paragraph}{\@startsection{paragraph}{4}{\z@}{0.5em \@plus 0.2em \@minus 0.1em}{-1em}{\normalfont\normalsize\bfseries}}
\makeatother

\usepackage{xspace}
\usepackage{booktabs}
\usepackage{multirow}
\usepackage{makecell}
\newcommand{\ours}{MAP-VP\xspace}%

\definecolor{cvprblue}{rgb}{0.21,0.49,0.74}
\usepackage[pagebackref,breaklinks,colorlinks,allcolors=cvprblue]{hyperref}

\def\arxivversion{}

\title{Reaching Is Not Finding: Map-Anchored Visual Prompting for the Last Mile of Aerial Vision-Language Navigation}

\author{%
Ningbo Li$^{1}$ \quad Kebin Liu$^{2,*}$ \quad Jingcan Chen$^{1}$ \quad Weichen Zhang$^{3}$\\
Rong Ding$^{1}$ \quad Yunhao Liu$^{2}$ \quad Mo Li$^{1,*}$\\[0.4em]
$^{1}$The Hong Kong University of Science and Technology \quad $^{2}$Tsinghua University\\
$^{3}$The University of Hong Kong\\[0.2em]
{\tt\small nlibc@connect.ust.hk \quad kebinliu2021@tsinghua.edu.cn \quad jchenhv@connect.ust.hk}\\
{\tt\small weic.zhang@connect.hku.hk \quad dingrong@ust.hk \quad yunhao@tsinghua.edu.cn}\\
{\tt\small lim@cse.ust.hk}\\[0.2em]
{\small $^{*}$Corresponding authors}
}

\begin{document}
\maketitle
\begin{strip}
\centering
\includegraphics[width=\textwidth]{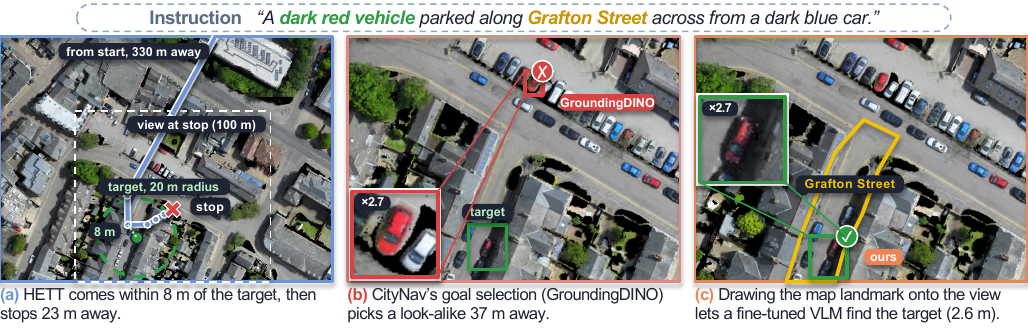}
\captionsetup{hypcap=false}
\captionof{figure}{\textbf{Reaching is not finding.} A real CityNav test episode. (a) The navigator (HETT) reaches the described car but stops outside the success radius. (b) At the same stop, CityNav's official goal-selection step, which detects the target phrase with GroundingDINO, picks a look-alike. (c) \ours, applied at the same stop, draws the map landmark onto the view and finds the target. Green: the described target; red: a wrong choice; yellow: the landmark.}
\captionsetup{hypcap=true}
\label{fig:\currfilebase}%
\end{strip}

\begin{abstract}
Aerial vision-language navigation (VLN) asks a drone to find an object described in natural language.
The description usually locates the object relative to named landmarks, such as streets and buildings.
We observe that current navigators often bring the drone close to the target and then stop at the wrong place: they reach the target without finding it.
This gap is shared by navigators of very different designs, from imitation learners to a fine-tuned 7B vision-language model.
A simple test reveals the cause.
When we replace the target object in the instruction with another one next to the same landmarks, no navigator goes to it, and most stop where they did before.
Navigators learn to reach the landmarks, not to identify the described object.
We therefore propose \ours, a plug-and-play module for the last step.
After any navigator stops, \ours draws the landmarks named in the instruction onto the drone's current view, using the landmark outlines that the benchmark provides on a map.
A fine-tuned 4B vision-language model then locates the target relative to these landmarks.
The training data are generated automatically from the benchmark annotations.
Neither drawing the landmarks nor fine-tuning helps much alone, but together they are highly effective.
Without retraining any navigator, \ours significantly improves all six navigators we test on CityNav.
It roughly doubles the success rate of the two strongest trained navigators, raising HETT from 25.9\% to 54.6\% and FlightGPT from 24.2\% to 48.9\%, whereas CityNav's official detector-based last step adds less than two points to HETT.
\end{abstract}

\section{Introduction}
\label{sec:intro}

Aerial vision-language navigation (VLN) asks a drone to find an object described in natural language, such as ``the white car parked on Staffordshire Street in front of Donegal''~\cite{liu2023aerialvln,lee2025citynav}.
CityNav~\cite{lee2025citynav} is a large-scale benchmark for this task built on real city scans.
An episode counts as a success only if the drone stops within 20\,m of the described object.
The task is hard because a city block contains many similar objects, and the description singles out one of them mainly through its relation to named landmarks.
Navigators of many designs have been proposed for CityNav, including imitation learners~\cite{anderson2018vln,krantz2020beyond}, map-based goal predictors~\cite{lee2024citynav}, two-stage transformers trained with reinforcement learning~\cite{ding2026hett,htnav2026}, and fine-tuned vision-language models (VLMs)~\cite{cai2025flightgpt}.
Yet in unseen environments their success rates remain far below that of human operators~\cite{lee2025citynav}.

We begin with the example in \cref{fig:teaser}, whose instruction asks for ``a dark red vehicle parked along Grafton Street across from a dark blue car''.
HETT~\cite{ding2026hett}, a state-of-the-art two-stage navigator, comes within 8\,m of this car but does not stop there.
It finally stops 23\,m away, outside the 20\,m success radius, so the episode counts as a failure (\cref{fig:teaser}a).
Yet the car is still visible in the drone's final top-down view, which covers about 100\,m$\times$100\,m.
Detecting the phrase ``dark red vehicle'' does not help either: the view contains many cars, and CityNav's official goal-selection step, which detects this phrase with GroundingDINO, picks a red car in a nearby parking lot (\cref{fig:teaser}b).
The navigator reached the target but did not find it.
This case is typical rather than rare.
In two thirds of HETT's failures on the test set, the drone stops outside the 20\,m success radius while the target is still in its final view.
The same holds for more than half of the failures of FlightGPT~\cite{cai2025flightgpt}, a fine-tuned 7B VLM.

To see whether this is a weakness of one particular navigator, we study six navigators of very different designs (\cref{sec:gap}).
Five of them are trained navigators with publicly released code and weights, covering the main design families on CityNav: end-to-end imitation learning, map-based goal prediction, a two-stage transformer with reinforcement learning, and a fine-tuned VLM.
The sixth is a training-free rule that flies straight to the landmarks named in the instruction.
Weak navigators often fail simply because they never get close to the target.
We therefore look only at the failures in which the target is still in the final view.
For these failures, all six navigators behave alike: a recognizer applied at the last step recovers 50--60\% of the failures of every navigator.
The last step is therefore a common bottleneck, independent of how the drone got there.
We call this shared bottleneck the \emph{reaching-is-not-finding} gap.

Why do navigators fail at the very last step?
We first report two observations and then discuss what they imply.
First, navigators learn to go to the landmarks: a training-free rule that flies to them outperforms every trained navigator we test.
Second, the last step ignores what the target is.
When we replace the target phrase in the instruction, for example ``building'' with ``parking lot'', HETT moves by only 4\,m although the new object lies about 40\,m away, and no other navigator reaches it either.

Together, these observations suggest that two things are missing at the last step.
The first is the right supervision.
Navigators are trained to imitate human trajectories, to regress goal coordinates, or to collect distance-based rewards; all of these signals reward getting close, but none rewards picking out the described instance.
The second is a spatial anchor.
The instruction locates the target relative to landmarks (``on Staffordshire Street, in front of Donegal''), yet at the last step nothing links these phrases to the pixels of the current view.
Detecting the target phrase alone does not fill this gap: CityNav's official goal-selection step with GroundingDINO~\cite{liu2024groundingdino} improves HETT by less than two points.
Recent works have also noticed that the final decision is weak~\cite{tong2026visa,xu2026sbfnav,wang2026raco}.
Their fixes, however, add only a few points and are built into a specific navigator.

In this paper, we close the last mile with \emph{map-anchored visual prompting}.
Our module, \ours, runs once after any navigator stops and draws the outlines and names of the landmarks mentioned in the instruction onto the drone's current top-down view.
CityNav already provides these landmarks as OpenStreetMap polygons, so the prompt adds no new information.
A 4B VLM~\cite{qwen3vl2025} then reads the image and the instruction and outputs a box around the target, and the drone flies to its center.
If the model reports that the target is not visible, the drone climbs and looks once more.
In the example of \cref{fig:teaser}c, drawing Grafton Street is enough for the model to pick the right car.
We fine-tune the VLM with LoRA~\cite{hu2022lora} on examples generated automatically from the training split, using the annotated object contours as boxes, so no human labeling is needed.
In terms of visual prompting~\cite{wu2024vpsurvey}, our marks are references rather than answers: they show named landmarks from a map, and the model has to infer where the target lies relative to them.

Our analysis (\cref{sec:analysis}) shows why such a simple module helps.
The visual anchors and the recognition supervision are each of little use alone, but strong together.
Without fine-tuning, the VLM treats any marked region as the answer, as reported for other open multimodal models~\cite{cai2024vipllava,yan2024somllava}.
After fine-tuning, the VLM uses the marks as references and locates the target relative to the drawn landmarks, so drawing the landmarks at wrong positions misleads the model.

Without retraining any navigator, \ours improves all six navigators we study.
The gain is largest for the two strongest trained navigators, whose success rates roughly double: HETT rises from 25.9\% to 54.6\%, and FlightGPT from 24.2\% to 48.9\%.
In summary, our contributions are:
\begin{itemize}
\item We identify the reaching-is-not-finding gap in aerial VLN, show that the gap is shared by six navigators of different designs, and trace the gap to the navigators' reliance on landmarks rather than on the target description.
\item We propose \ours, a plug-and-play module that combines map-anchored visual prompts with automatically generated recognition supervision.
\item We show large gains on all six navigators and explain the gains through controlled analyses of the anchors, the supervision, and the model's attention.
\end{itemize}

\section{Related Work}
\label{sec:related}

\paragraph{Aerial vision-language navigation.}
AerialVLN~\cite{liu2023aerialvln} and CityNav~\cite{lee2025citynav} introduced language-guided drone navigation in large outdoor scenes.
CityNav builds on the object descriptions of CityRefer~\cite{miyanishi2023cityrefer} and provides the OpenStreetMap polygons of the landmarks mentioned in each description.
Trained navigators on CityNav include the map-based goal predictor (MGP) of CityNav, the history-enhanced two-stage transformer HETT~\cite{ding2026hett}, and the hybrid imitation and reinforcement learner HTNav~\cite{htnav2026}.
FlightGPT~\cite{cai2025flightgpt} fine-tunes Qwen2.5-VL on a semantic map in which the landmarks are painted as masks.
Zero-shot agents build on large multimodal models, \eg GeoNav~\cite{xu2026geonav}, CLOSER-VLN~\cite{li2026closer}, and ViSA~\cite{tong2026visa}, while SBFNav~\cite{xu2026sbfnav} keeps several goal hypotheses in a spatial belief field.

Several of these works point to the final decision as a weakness.
ViSA attributes failures of earlier agents to missing confirmation and verifies numbered detection candidates in three stages.
SBFNav reports a large gap between its best candidate and the one its selector picks, and RACO~\cite{wang2026raco} corrects unreliable coarse goals of a reproduced HETT.
On TravelUAV, See-and-Reach~\cite{xue2026seereach} studies the approach phase once the target is in view.
Our work differs in three ways.
We measure the gap across six navigators, we identify its cause by perturbing the instruction, and we close it with a module that does not depend on the navigator.
We also note that the illustrations of ViSA overlay landmark outlines on the image.
Its method, however, is built around numbered detection candidates and neither describes nor ablates this overlay.
In \ours, the map-anchored landmarks are the only prompt, and we isolate their effect.

\paragraph{Verification and stopping in embodied navigation.}
Deciding where to stop is a known weakness of navigation agents.
ConsistNav~\cite{wang2026consistnav} adds a training-free execution layer to object-goal navigation that suppresses stops without verification, and CLOSER-VLN verifies each action with a fine-tuned VLM.
\ours instead acts once, after the navigator has finished.
It treats the last step as referring-expression comprehension in drone imagery~\cite{sun2025refdrone}, \ie as picking out the described instance among many similar ones.

\paragraph{Visual prompting for multimodal models.}
The term visual prompting covers two different ideas: learning prompt tokens~\cite{jia2022vpt}, or drawing marks on the image for the model to read~\cite{wu2024vpsurvey}.
We follow the second.
Marks such as red circles~\cite{shtedritski2023redcircle}, Set-of-Mark labels~\cite{yang2023som}, and coordinate grids~\cite{lei2024scaffold} work well without training on strong proprietary models.
Open multimodal models, in contrast, need instruction tuning before they can read arbitrary marks~\cite{cai2024vipllava,yan2024somllava,lin2025drawunderstand}.
In embodied tasks, marks usually enumerate candidate actions or targets~\cite{nasiriany2024pivot,sathyamoorthy2024convoi}, and in remote sensing, EarthMarker~\cite{zhang2024earthmarker} marks the region being asked about.
In all of these works, the marked region is either the object of the question or a candidate answer.
VTPrompt~\cite{jiang2024vtprompt}, whose pipeline is closest to ours, also highlights detected answer objects.
In \ours, the marks are named landmarks projected from an external map, and the answer is somewhere else, to be inferred from the spatial relations in the instruction.
We find that this reference role is exactly what zero-shot models get wrong and what fine-tuning teaches (\cref{sec:analysis}).

\section{Reaching Is Not Finding}
\label{sec:gap}
Here we measure this gap and trace its cause.

\subsection{Setting}
\label{sec:setting}
\paragraph{Benchmark.}
CityNav~\cite{lee2025citynav} contains about 32k episodes over 34 urban blocks of Birmingham and Cambridge, scanned in SensatUrban~\cite{hu2021sensaturban}.
Each episode provides a free-form description of a target object, which is a building, a car, a parking lot, or a ground area.
It also provides the names of the landmarks mentioned in the description, their OpenStreetMap polygons, and a start pose about 50\,m above the ground.
The drone observes a top-down crop of an orthophoto, rendered at 0.1\,m per pixel around its current pose.
The simulator gives this pose in the world frame and the orthophoto is geo-referenced, so every pixel has known world coordinates, as do the landmark polygons and the goals predicted by MGP and HETT.
We use the revised annotations of HETT~\cite{ding2026hett}, which correct the landmark lists and drop a few episodes (\cref{sec:addres}); our module is trained with them. We also report the original annotations.

\paragraph{Metrics and splits.}
Following the code of MGP, HETT, FlightGPT, and GeoNav, an episode succeeds if the final position lies within 20\,m of the target in the horizontal plane (SR).
We also report SPL and the success rate at 10\,m (SR@10).
We evaluate on the \textit{test-unseen} split, with 5,281 revised episodes (5,311 originally) on six unseen blocks.
We also use a held-out validation set of 4,867 episodes from \textit{val-seen} and \textit{val-unseen}, after setting aside 300 episodes for tuning.
All design choices are made on the validation set.

\paragraph{Navigators.}
We study six navigators of different designs.
Seq2Seq~\cite{anderson2018vln} and CMA~\cite{krantz2020beyond} are end-to-end imitation learners provided as CityNav baselines.
MGP~\cite{lee2024citynav} predicts the goal on a map, FlightGPT~\cite{cai2025flightgpt} is a 7B VLM trained with SFT and GRPO, and HETT~\cite{ding2026hett} is a two-stage transformer trained with reinforcement learning.
The sixth, which we call \emph{landmark-centroid}, is a training-free rule that flies to the mean of the vertices of all landmark polygons named in the instruction, with the official controller that MGP uses to reach its goals.
We choose trained navigators whose code and trained weights are publicly available, so that their end points can be reproduced exactly, and that together cover the main design families.
For the CityNav baselines we use the weights trained on human demonstrations.
Methods without released weights, such as HTNav, CLOSER-VLN, SBFNav, and ViSA, are compared through their reported numbers (\cref{tab:sota}).
With the official inference code, our results are close to the published ones for FlightGPT (24.24 \vs 24.47), Seq2Seq, and CMA.
HETT reaches 25.94 \vs 28.90 reported, which matches an independent reproduction (25.73~\cite{wang2026raco}) and the authors' note that the released checkpoint is not the one used in their paper.

\subsection{Failures happen near the target}
\label{sec:decomp}
\begin{figure}[t]
\centering
\includegraphics[width=\linewidth]{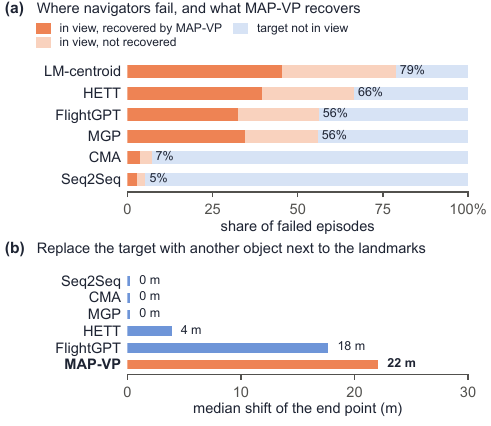}
\caption{\textbf{Reaching is not finding} (test-unseen). (a)~Failed episodes of each navigator by whether the target is inside the view from 50\,m at the stop, and the share that \ours turns into successes. (b)~Median shift of the end point when the target in the instruction is replaced by another object next to the same landmarks (386 episodes; the new object lies a median of 40\,m away). \ours is applied at HETT's end points. Exact numbers are in the supplementary material (\cref{tab:decomp,tab:perturb}).}
\label{fig:gap}
\end{figure}
Many failures end close to the target.
To show this, \cref{fig:gap}a groups the failed episodes of each navigator by where the drone stops.
A failed episode by definition ends more than 20\,m from the target.
We say that the target is \emph{in view} if it lies inside the 100\,m$\times$100\,m area that the drone sees from 50\,m above its final position.

For the navigators that usually reach the vicinity of the target, namely FlightGPT, HETT, and landmark-centroid, more than half of the failures end within 50\,m of the target.
In 56--79\% of their failures, the target is in view.
HETT and FlightGPT also pass within 20\,m of the target and still stop outside in 21.5\% and 17.3\% of all episodes, respectively.
Seq2Seq and CMA, in contrast, fail mainly because they never get close; MGP lies in between.

The darkest part of each bar in \cref{fig:gap}a gives the most important result.
It applies our last-step module (\cref{sec:method}) to the failures in which the target is in view.
The module recovers 50--62\% of these failures for every navigator.
The last step is therefore a shared bottleneck that does not depend on how the drone got there.

\begin{figure*}[t]
\centering
\includegraphics[width=\textwidth]{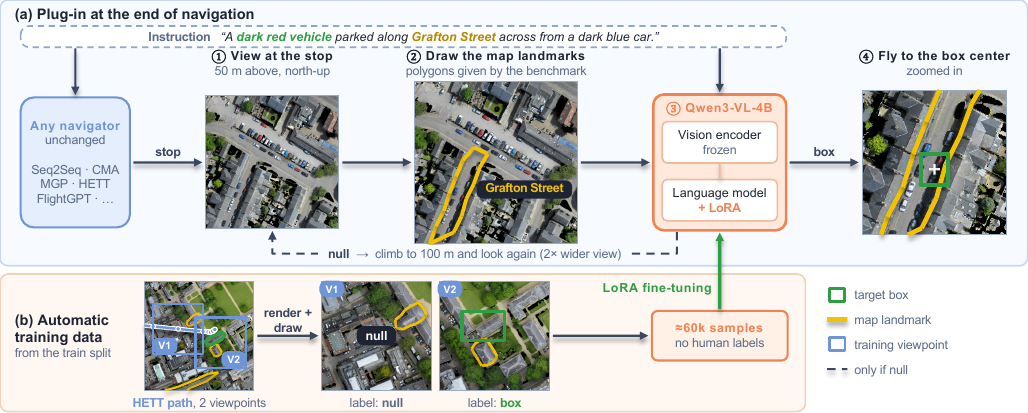}
\caption{\textbf{Overview of \ours.} (a)~After any navigator stops, \ours draws the instruction's map landmarks onto the view (1--2); a fine-tuned VLM returns a box or \texttt{null} (3); the drone flies to the box center (4), or climbs to 100\,m and looks again. (b)~Training views from the train split, at HETT's decision points (V1) and at random points near the target (V2), labeled with the target box or \texttt{null}.}
\label{fig:method}
\end{figure*}

\subsection{Navigators reach landmarks, not targets}
\label{sec:why}
We now examine why the navigators fail at the last step even when the target is in view.
We first show that they learn to go to the landmarks, and then test directly whether they use the target description at the end.

\paragraph{Navigators learn to go to the landmarks.}
HETT's final positions lie a median of 24\,m from the centroid of the landmarks named in the instruction.
On a 300-episode validation subset, the error of its first-stage goal (median 36\,m) is about the same as the error of that centroid itself (35\,m): HETT predicts roughly where the landmarks are.
This prior alone goes far: a training-free rule that flies to the landmark centroid reaches 40.2\% SR on test-unseen (30.6\% with the original annotations), higher than any trained navigator we study.

\paragraph{The last step ignores the target description.}
To test whether the navigators use the target description at the end, we replace the target in the instruction and keep everything else fixed (\cref{fig:gap}b).
We replace its first mention with the target phrase of an object of another category on the same map (\eg ``building'' $\rightarrow$ ``parking lot'').
A navigator that uses the target should then fly to an object of the new category next to the same landmarks.
We keep the 386 test episodes with exactly one such object, more than 20\,m from the original target (median 40\,m; \cref{sec:swap}).

Seq2Seq, CMA, and MGP stop at the same place in most of these episodes, and HETT moves by a median of only 4\,m: they go where the landmarks are and ignore what the target is.
FlightGPT, which reads the full instruction with a 7B VLM, moves by 18\,m.
Its success rate for the new object (final position within 20\,m) is 15.8\% with the original instruction and still 15.3\% with the swapped one: it moves, but not towards the new object.
\ours at HETT's end points moves by a similar 22\,m, and its success rate rises from 10.1\% to 30.6\%: it moves towards the new object.

\paragraph{Detecting the target phrase is not enough.}
CityNav also has an official goal-selection step that runs after navigation.
It detects the target phrase in the final view with GroundingDINO~\cite{liu2024groundingdino} and flies to the highest-scoring box.
On top of HETT, this step adds only 1.9 points (25.94 $\rightarrow$ 27.84).
At the goals predicted by MGP, it adds 0.5 points (21.95 $\rightarrow$ 22.44, \cref{sec:mgp}).
A phrase such as ``white car'' matches many objects in the view.
The detector score also barely tells whether the described object is visible at all (AUROC 0.52 on 400 validation views).

\paragraph{Summary.}
Trained with signals that reward getting close, the navigators learn to go where the landmarks are.
At the final step they lack both supervision for identifying the described instance and an anchor that ties the spatial relations in the instruction to the pixels of the current view.

\section{Map-Anchored Visual Prompting}
\label{sec:method}

We add the two ingredients missing in \cref{sec:gap}, a spatial anchor and recognition supervision, without changing the navigator.
\ours is therefore a separate module that runs once, after any navigator has stopped (\cref{fig:method}).

\subsection{Observation at the end of navigation}
\label{sec:obs}
At the final position, the drone takes a top-down view from $h{=}50$\,m above the ground.
The view is oriented north-up and covers $2h\times2h$\,m at $10h$ pixels, \ie 0.1\,m per pixel.
This is the same observation that CityNav's official goal-selection step uses~\cite{lee2024citynav}, and it is cropped by the same function that produces the navigators' own observations.
\ours only looks at the area below the drone's current position and never accesses views of places the drone has not visited.

\subsection{Map-anchored visual prompts}
\label{sec:prompt}
CityNav provides the names of the landmarks mentioned in each instruction, together with their OpenStreetMap polygons in the world frame.
MGP and HETT rasterize these polygons into a map input, and FlightGPT paints them as masks on an overhead image of the whole area.
Because the view is geo-referenced (\cref{sec:setting}), we can project each polygon exactly into the current image, using only the drone's current pose.
We draw its outline in yellow and write its name next to the visible part.
If a landmark lies outside the image, an arrow at the image border points towards it.
The prompt adds no information beyond what the navigators already receive, and it never marks the target (\cref{sec:fair}).
What it changes is the form of the information.
Spatial relations in the instruction, such as ``on Staffordshire Street'' or ``in front of Donegal'', become geometry that the model can see in the image.

\subsection{Recognition supervision from benchmark annotations}
\label{sec:data}
To teach a VLM to use these prompts, we generate training data automatically from the \emph{train-seen} split.
Each sample is a view from a chosen point.
Half of the points are decision points actually visited by HETT on training episodes, so the data reflects where navigators really stop, often with the target out of view.
The other half are random points within 45\,m of the target.
Half of the views are north-up and half are aligned with the drone's heading.
We draw the landmark prompts as above and project the annotated contour of the target into the image.
The bounding box of this contour is the answer.
If the target lies outside the image, the answer is \texttt{null}, meaning ``not visible''.
The model outputs the box as JSON in normalized coordinates.
No human labeling or external model is involved, and generating 60k samples takes a few minutes.

\subsection{Model and training}
\label{sec:model}
We fine-tune Qwen3-VL-4B-Instruct~\cite{qwen3vl2025} with LoRA~\cite{hu2022lora} on all linear layers of the language model (33M trainable parameters, under 1\% of the model) for one epoch on the 60k samples (1.5--3 hours on two GPUs); one forward pass takes about 0.8\,s (other hyper-parameters in \cref{sec:impl}).

\subsection{Acting on the answer and looking again}
\label{sec:act}
If the model outputs a box, the drone flies to the corresponding ground point with the navigator's own actions (5\,m forward or 30$^\circ$ turns) and stops there.
\ours may use 20 actions and never lengthens the episode: if the navigator has fewer actions left, \ours takes over its last steps (\cref{sec:budget}).
The flight is also added to the path length for SPL.
If the model answers \texttt{null}, the target is most likely outside the view from 50\,m.
In that case the drone climbs to 100\,m, which takes one more step, and looks once more, using the same model on a 200\,m$\times$200\,m view.
We render this view at 1000\,px, which keeps the ground resolution at 0.1\,m per pixel.
The 100\,m of vertical flight is added to the path length.
We call this step \emph{look-again} (LA).
Note that the model has only seen views from 50\,m during training.

\section{Experiments}
\label{sec:exp}

\begin{table*}[t]
\centering
\small
\setlength{\tabcolsep}{4pt}
\resizebox{\textwidth}{!}{%
\begin{tabular}{ll ccc ccc ccc c}
\toprule
 & & \multicolumn{3}{c}{Navigator} & \multicolumn{3}{c}{+ \ours} & \multicolumn{3}{c}{+ \ours + LA} & Val SR \\
\cmidrule(lr){3-5}\cmidrule(lr){6-8}\cmidrule(lr){9-11}\cmidrule(lr){12-12}
Navigator & Type & SR & SPL & SR@10 & SR & SPL & SR@10 & SR & SPL & SR@10 & nav $\rightarrow$ ours+LA \\
\midrule
Seq2Seq~\cite{anderson2018vln}$^{\dagger\ddagger}$ & imitation, end-to-end & 1.19 & 0.72 & 0.33 & 3.87 & 2.28 & 2.85 & 6.98 & 3.65 & 4.67 & 0.84 $\rightarrow$ 6.16 \\
CMA~\cite{krantz2020beyond}$^{\dagger\ddagger}$ & imitation, end-to-end & 1.04 & 0.49 & 0.29 & 4.61 & 2.18 & 3.09 & 9.19 & 3.85 & 5.84 & 0.84 $\rightarrow$ 7.84 \\
MGP$^{\dagger\S}$~\cite{lee2024citynav} & map-based goal prediction & 8.46 & 8.01 & 2.06 & 41.56 & 34.72 & 28.40 & 48.31 & 38.87 & 32.61 & 7.91 $\rightarrow$ 44.69 \\
FlightGPT$^\dagger$~\cite{cai2025flightgpt} & 7B VLM, SFT + GRPO & 24.24 & 21.95 & 7.48 & 45.05 & 36.51 & 31.47 & 48.85 & 38.18 & 33.91 & 16.52 $\rightarrow$ 43.19 \\
HETT$^\dagger$~\cite{ding2026hett} & two-stage, IL + RL & 25.94 & 21.17 & 10.09 & 51.56 & 38.31 & 36.38 & 54.55 & 39.96 & 37.99 & 23.40 $\rightarrow$ 51.65 \\
LM-centroid & training-free & 40.20 & 38.77 & 13.37 & 59.55 & 52.64 & 42.85 & \textbf{61.37} & \textbf{53.49} & \textbf{44.14} & 24.61 $\rightarrow$ 52.00 \\
\bottomrule
\end{tabular}}
\caption{\textbf{\ours on six navigators} (CityNav test-unseen, 5,281 episodes with the revised annotations of HETT~\cite{ding2026hett}). $^\dagger$Reproduced by us with the released code and weights. $^\ddagger$5,117 episodes kept by the official VLN-CE loader; Seq2Seq and CMA do not use the landmark polygons that \ours draws (\cref{sec:fair}). $^\S$With CityNav's official goal-selection step; \ours replaces this step and starts where the drone stops (\cref{sec:mgp}).
LA: look again from 100\,m when the view from 50\,m gives ``not visible''. \ours stays within each navigator's step limit (\cref{sec:budget}), and its flight counts in SPL.
Every gain is significant: the paired bootstrap 95\% confidence interval of the SR gain excludes zero in every row, \eg HETT +25.6 [+24.2, +27.0].
Val SR: held-out validation set (4,867 episodes).}
\label{tab:main}
\end{table*}

\subsection{Main results}
\label{sec:main}
\Cref{tab:main} shows that \ours improves every navigator by a large margin on both test and validation.
The gains are even larger at the stricter 10\,m threshold; for HETT, SR@10 rises from 10.1\% to 38.0\%.
The size of the gain depends on how often a navigator brings the target into view.
For HETT and FlightGPT, \ours roughly doubles the success rate.
Seq2Seq and CMA, in contrast, stop more than 75\,m from the target in 87--90\% of the episodes, so the module can only help the few episodes that end nearby.
For these navigators, looking again from 100\,m matters most; for CMA it raises SR from 4.6\% to 9.2\%.
For HETT, look-again is triggered in only 7\% of the episodes and improves both SR and SPL.
For MGP, 5.2 points of the gain come from skipping its goal selection (\cref{sec:mgp}).

\subsection{Comparison with published results}
\label{sec:compare}
\Cref{tab:sota} lists the reported results of published methods next to ours, separately for the revised and the original annotations.
With the revised annotations, HETT + \ours + LA reaches 54.6\% and LM-centroid + \ours + LA reaches 61.4\%.
The highest result among the compared methods is 36.1\%, obtained by ViSA~\cite{tong2026visa} with several calls to a proprietary VLM at each step.
With the original annotations, LM-centroid + \ours + LA reaches 49.4\%, compared with 32.3\% for SBFNav~\cite{xu2026sbfnav}.
LM-centroid alone (40.2\% and 30.6\%) already exceeds most published navigators.
We report it as a baseline, not as our method.

\begin{table}[t]
\centering
\small
\setlength{\tabcolsep}{4pt}
\begin{tabular}{lcc}
\toprule
Method & Revised & Original \\
\midrule
MGP~\cite{lee2024citynav} & \phantom{0}9.70 & \phantom{0}6.38 \\
FlightGPT~\cite{cai2025flightgpt} & 24.47 & 21.20 \\
HTNav~\cite{htnav2026} & 25.49 & 22.23 \\
HETT~\cite{ding2026hett} & 28.90 & 22.97 \\
GeoNav~\cite{xu2026geonav} & -- & 25.9\phantom{0} \\
CLOSER-VLN~\cite{li2026closer} & -- & 32.01 \\
SBFNav~\cite{xu2026sbfnav} & 32.46 & 32.29 \\
ViSA~\cite{tong2026visa} & 36.11 & -- \\
\midrule
HETT$^\dagger$ + \ours + LA & 54.55 & -- \\
MGP$^\dagger$ + \ours + LA & 48.31 & 38.99 \\
LM-centroid + \ours + LA & \textbf{61.37} & \textbf{49.43} \\
\bottomrule
\end{tabular}
\caption{\textbf{Comparison with reported results} (test-unseen SR, \%). Revised: annotations of HETT~\cite{ding2026hett}. Numbers of other methods are as reported; the revised numbers of MGP and FlightGPT are taken from~\cite{htnav2026}. Our original results use all 5,311 episodes. $^\dagger$Reproduced by us with the released code and weights. HETT's weights are trained on the revised annotations only.}
\label{tab:sota}
\end{table}

\subsection{Same viewpoint, different selector}
\label{sec:selector}
A fair comparison of last-step selectors should remove any effect of navigation.
We therefore run them at exactly the same viewpoints: the goals predicted by MGP, where CityNav's official GroundingDINO goal selection operates (\cref{tab:selector} in the supplementary material).
At these viewpoints, GroundingDINO adds 0.5 points and \ours adds 27.1.
We fix a coordinate error in the GroundingDINO step, which as released lowers SR to 8.5\% (\cref{sec:mgp}).

\subsection{The gain depends on where the navigator stops}
\label{sec:distance}
\begin{figure}[t]
\centering
\includegraphics[width=\linewidth]{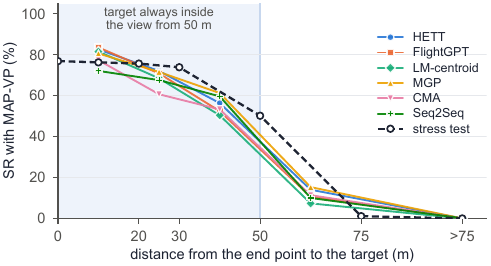}
\caption{\textbf{Success of \ours \vs the navigator's end-point distance} (test-unseen, view from 50\,m only). Points are distance bins; the last one pools all end points beyond 75\,m. All six navigators follow one curve, and so does a stress test that places the end point at a set distance from the target in a random direction (dashed, 1,000 episodes per distance). Shaded: the target is always inside the view.}
\label{fig:distance}
\end{figure}

\Cref{fig:distance} plots the success rate of \ours against the distance between the navigator's end point and the target.
The six navigators fall on almost the same curve.
\ours succeeds in about 80\% of the episodes that end within 20\,m, 70\% at 20--30\,m, and 55\% at 30--50\,m.
The rate drops to 7--15\% at 50--75\,m and to zero beyond 75\,m.
A stress test gives the same result: when we place the end point at a controlled distance from the target in a random direction, \ours succeeds in 74--77\% of the cases as long as the target is inside the view from 50\,m.
The gain on a new navigator can therefore be predicted from its end-point distances.

\subsection{Why does map-anchored prompting work?}
\label{sec:analysis}
All analyses use HETT's end points on test-unseen and fly straight to the predicted point; all fine-tuned models use the same 60k samples, and all models get the same prompt.

\paragraph{Anchors and recognition supervision are both needed.}
\begin{figure}[t]
\centering
\includegraphics[width=\linewidth]{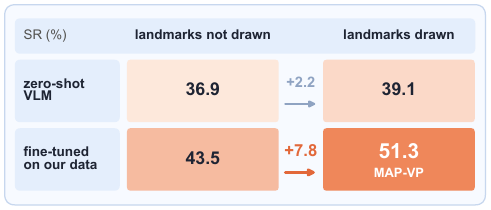}
\caption{\textbf{Anchors $\times$ recognition supervision.} SR (\%) of the last step at HETT's end points (test-unseen). Drawing the landmarks helps a zero-shot VLM little (+2.2) but helps a lot after fine-tuning on our data (+7.8). Validation results are in \cref{tab:ladder} (supplementary material).}
\label{fig:interaction}
\end{figure}
\Cref{fig:interaction} crosses two factors: whether the landmarks are drawn into the view, and whether the VLM is fine-tuned on our recognition data.
Reading the full instruction with a zero-shot VLM already helps, raising SR from 27.8\% with GroundingDINO to 36.9\%.
Drawing the landmarks for the zero-shot model, however, changes little (39.1\%), and fine-tuning without drawn landmarks reaches only 43.5\%.
Combining the two raises SR to 51.3\%.
The interaction, \ie the gain from fine-tuning with landmarks minus the gain from fine-tuning without them, is +5.6 points [+4.0, +7.2].

Scaling the model or the input does not replace either ingredient.
An 8B model is only 0.5 points better than the 4B model trained on the same samples (not significant).
Raising the input resolution from 224 to 1000\,px improves localization within 20\,m by only 4 points on a probe set.
The bottleneck is relating the description to the right instance, not seeing the image.

\paragraph{Zero-shot models treat marks as answers.}
The full results (\cref{tab:ladder}) also hint at why the drawn landmarks do not help without fine-tuning.
Once the landmarks are drawn, the zero-shot VLM places its box on a landmark more often (17.4\% $\rightarrow$ 23.1\%).
We test this behavior directly by adding one unnamed yellow box to the view (\cref{tab:marks}, right).
For the zero-shot model, the answer follows the box.
A box around the true target raises SR from 46.7\% to 65.6\%, and a box around a distractor of the same category lowers it from 38.2\% to 31.2\%.
After fine-tuning, neither box has any effect.

This behavior matches the convention of most visual prompting data, in which the marked region is what the question is about~\cite{cai2024vipllava,yan2024somllava}.
Our recognition supervision teaches the model that the marks in our setting are references instead.
Simply telling the zero-shot model that the yellow outlines are references and not the target does not achieve this.
With such a sentence in the prompt, SR rises by only 0.8 points, and the share of boxes on landmarks grows to 24.3\%.

\begin{table}[t]
\centering
\small
\setlength{\tabcolsep}{3.5pt}
\resizebox{\linewidth}{!}{%
\begin{tabular}{l cc c cc}
\toprule
 & \multicolumn{2}{c}{Landmark positions} & & \multicolumn{2}{c}{Extra unnamed box} \\
\cmidrule(lr){2-3}\cmidrule(lr){5-6}
Model & correct & shifted 40\,m & & on target & on distractor \\
\midrule
Zero-shot & 39.06 & -- & & 46.7 $\rightarrow$ \textbf{65.6} & 38.2 $\rightarrow$ \textbf{31.2} \\
\ours & 51.30 & 34.24 & & 61.2 $\rightarrow$ 61.4 & 50.8 $\rightarrow$ 50.1 \\
\bottomrule
\end{tabular}}
\caption{\textbf{Does the model use the drawn landmarks, and how?} Left: SR when all drawn landmarks are shifted by 40\,m; for reference, the model fine-tuned without drawn landmarks reaches 43.53. Right: SR without $\rightarrow$ with one extra unnamed box, on the subsets where a target box (4,404 episodes) or a distractor box (5,081 episodes) can be drawn.}
\label{tab:marks}
\end{table}

\paragraph{The fine-tuned model locates the target relative to the landmarks.}
\begin{figure}[t]
\centering
\includegraphics[width=\linewidth]{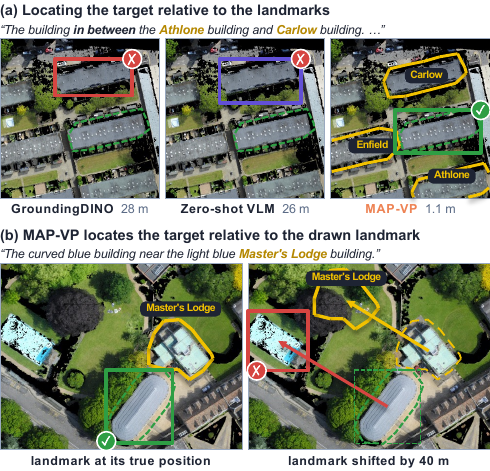}
\caption{\textbf{Qualitative examples} (test-unseen). Dashed green: the target. (a)~Distances from the chosen box to the target. (b)~Only the drawn outline is shifted by 40\,m (image, instruction, and visible target unchanged); the box of \ours moves with it.}
\label{fig:qual}
\end{figure}
If the fine-tuned model used the drawn landmarks only as generic context, moving them would cost little.
Instead, shifting all landmark outlines by 40\,m in a random direction lowers SR from 51.3\% to 34.2\% (\cref{tab:marks}, left), and the box moves with the landmark (\cref{fig:qual}b).
The result is even below that of the model fine-tuned without any drawn landmarks.
Swapping the names among the outlines, in episodes with at least two landmarks, lowers SR by 6.1 points.
The model thus reads where the landmarks are and which is which, and infers the target position from them.

\paragraph{Attention: first the landmark, then the target.}
We also measure where the model attends in the image.
On 1,000 test end points with the target in view, we average the attention over heads and over the middle layers 9--24 (of 36, from 0), normalized over the visual tokens.
We fixed these layers in advance: question-relevant visual information is retrieved in the middle layers~\cite{zhang2025crossmodal}, and averaging layers is a suitable default when no layer is tuned~\cite{zhang2025mllmsknow}.
When the outlines are drawn, the tokens of a landmark name put 27\% of their visual attention inside that landmark's outline.
Without the drawing, the share is 13\%, while the outlines cover 11\% of the image.
This holds with and without fine-tuning: the drawing alone grounds the landmark name in the image.

At the position where the model starts its answer, the picture changes with fine-tuning.
With the landmarks drawn, the described target is the most attended of the similar objects in the view in 24\% of the views for the zero-shot model and in 34\% for the fine-tuned model.
Attention on the target also tells whether the answer will be correct: in a pair of one success and one failure, the success attends more to its target in 81\% of pairs for the fine-tuned model and in 68\% for the zero-shot model (AUC; details in \cref{sec:impl}).
Fine-tuning thus focuses the model on the described instance.

\paragraph{What does not help.}
Numbered candidates in the style of Set-of-Mark, coarse-to-fine re-asking, and stopping during flight do not beat a single prediction at the end point (\cref{sec:nothelp}).

\paragraph{What ``not visible'' means.}
The fine-tuned model answers \texttt{null} at 7.1\% of HETT's end points.
In 68\% of these cases, the target is indeed outside the view from 50\,m, and in 90\% it is more than 50\,m away, so looking again from 100\,m addresses the right cases.
Among the episodes whose target is outside the view from 50\,m, a single look from 100\,m raises SR from 4.9\% to 34.2\%.

\section{Limitations and Conclusion}
\label{sec:conclusion}

\paragraph{Limitations.}
\ours repairs the last step, not navigation itself.
When a navigator stops more than about 75\,m from the target, the target is outside the view from 50\,m (\cref{fig:distance}), and looking again from 100\,m recovers only a small share of these episodes.
Looking again from 100\,m uses a 1000\,px view to keep 0.1\,m per pixel, which a real camera would need to match.
Like CityNav's official goal-selection step, \ours acts after the navigator has used its step budget.
All of its flight is charged to SPL, but we have not evaluated a variant that reserves part of the budget for it.
Like the CityNav benchmark, we assume an accurate pose and landmark map; real GPS and map errors would offset the outlines, and we test only a 40\,m shift.
The prompts also depend on the quality of the landmark lists.
With the original annotations, 437 of the 5,311 test episodes have no matched landmark, and \ours cannot anchor them.
When the target is in view, most remaining errors pick a similar object near the right landmark, and descriptions that require counting are harder (\cref{sec:fail}).
Finally, our experiments are limited to CityNav, whose instructions all come with map-registered landmarks.

\paragraph{Conclusion.}
Aerial VLN navigators of very different designs reach the vicinity of the described target but often fail to identify it, because they learn where the landmarks are and ignore what the target is.
We close this last mile with a plug-and-play module that draws the map landmarks of the instruction onto the current view and learns from automatically generated supervision to locate the target relative to them.
Without retraining any navigator, it roughly doubles the success rate of the two strongest trained navigators.

\section*{LLM Usage Statement}
We used Claude (Anthropic) through the Claude Code assistant to draft and edit the text, to write and debug the code for experiments and figures, and to search related work.
The authors checked all text, code, and citations, every reported number comes from running this code, and the authors take full responsibility for the content.

{
    \small
    \bibliographystyle{ieeenat_fullname}
    \bibliography{main}

\begin{thebibliography}{35}
\providecommand{\natexlab}[1]{#1}
\providecommand{\url}[1]{\texttt{#1}}
\expandafter\ifx\csname urlstyle\endcsname\relax
  \providecommand{\doi}[1]{doi: #1}\else
  \providecommand{\doi}{doi: \begingroup \urlstyle{rm}\Url}\fi

\bibitem[Anderson et~al.(2018)Anderson, Wu, Teney, Bruce, Johnson,
  S{\"u}nderhauf, Reid, Gould, and van~den Hengel]{anderson2018vln}
Peter Anderson, Qi Wu, Damien Teney, Jake Bruce, Mark Johnson, Niko
  S{\"u}nderhauf, Ian Reid, Stephen Gould, and Anton van~den Hengel.
\newblock Vision-and-language navigation: Interpreting visually-grounded
  navigation instructions in real environments.
\newblock In \emph{CVPR}, 2018.

\bibitem[Bai et~al.(2025)Bai, Cai, Chen, Chen, Chen, Cheng, Deng, Ding, Gao,
  Ge, et~al.]{qwen3vl2025}
Shuai Bai, Yuxuan Cai, Ruizhe Chen, Keqin Chen, Xionghui Chen, Zesen Cheng,
  Lianghao Deng, Wei Ding, Chang Gao, Chunjiang Ge, et~al.
\newblock {Qwen3-VL} technical report.
\newblock \emph{arXiv preprint arXiv:2511.21631}, 2025.

\bibitem[Cai et~al.(2025)Cai, Dong, Tan, Deng, Li, Gao, Wang, Su, Sumalee, and
  Zhong]{cai2025flightgpt}
Hengxing Cai, Jinhan Dong, Jingjun Tan, Jingcheng Deng, Sihang Li, Zhifeng Gao,
  Haidong Wang, Zicheng Su, Agachai Sumalee, and Renxin Zhong.
\newblock {FlightGPT}: Towards generalizable and interpretable {UAV}
  vision-and-language navigation with vision-language models.
\newblock In \emph{EMNLP}, 2025.

\bibitem[Cai et~al.(2024)Cai, Liu, Mustikovela, Meyer, Chai, Park, and
  Lee]{cai2024vipllava}
Mu Cai, Haotian Liu, Siva~Karthik Mustikovela, Gregory~P. Meyer, Yuning Chai,
  Dennis Park, and Yong~Jae Lee.
\newblock {ViP-LLaVA}: Making large multimodal models understand arbitrary
  visual prompts.
\newblock In \emph{CVPR}, 2024.

\bibitem[Ding et~al.(2026)Ding, Gao, Pan, Wang, and Qin]{ding2026hett}
Xichen Ding, Jianzhe Gao, Cong Pan, Wenguan Wang, and Jie Qin.
\newblock History-enhanced two-stage transformer for aerial vision-and-language
  navigation.
\newblock In \emph{AAAI}, 2026.

\bibitem[Fan et~al.(2026)Fan, Pan, Liu, Liu, and Qin]{htnav2026}
Chengjie Fan, Cong Pan, Zijian Liu, Ningzhong Liu, and Jie Qin.
\newblock {HTNav}: A hybrid navigation framework with tiered structure for
  urban aerial vision-and-language navigation.
\newblock In \emph{CVPR}, 2026.

\bibitem[Hu et~al.(2022)Hu, Shen, Wallis, Allen-Zhu, Li, Wang, Wang, and
  Chen]{hu2022lora}
Edward~J. Hu, Yelong Shen, Phillip Wallis, Zeyuan Allen-Zhu, Yuanzhi Li, Shean
  Wang, Lu Wang, and Weizhu Chen.
\newblock {LoRA}: Low-rank adaptation of large language models.
\newblock In \emph{ICLR}, 2022.

\bibitem[Hu et~al.(2021)Hu, Yang, Khalid, Xiao, Trigoni, and
  Markham]{hu2021sensaturban}
Qingyong Hu, Bo Yang, Sheikh Khalid, Wen Xiao, Niki Trigoni, and Andrew
  Markham.
\newblock Towards semantic segmentation of urban-scale {3D} point clouds: A
  dataset, benchmarks and challenges.
\newblock In \emph{CVPR}, 2021.

\bibitem[Jia et~al.(2022)Jia, Tang, Chen, Cardie, Belongie, Hariharan, and
  Lim]{jia2022vpt}
Menglin Jia, Luming Tang, Bor-Chun Chen, Claire Cardie, Serge Belongie, Bharath
  Hariharan, and Ser-Nam Lim.
\newblock Visual prompt tuning.
\newblock In \emph{ECCV}, 2022.

\bibitem[Jiang et~al.(2025)Jiang, Zhang, Zhou, Jin, and Liu]{jiang2024vtprompt}
Songtao Jiang, Yan Zhang, Chenyi Zhou, Yeying Jin, and Zuozhu Liu.
\newblock Joint visual and text prompting for zero-shot object-oriented
  perception with multimodal large language models.
\newblock In \emph{Advanced Hybrid Information Processing}. Springer, 2025.

\bibitem[Krantz et~al.(2020)Krantz, Wijmans, Majumdar, Batra, and
  Lee]{krantz2020beyond}
Jacob Krantz, Erik Wijmans, Arjun Majumdar, Dhruv Batra, and Stefan Lee.
\newblock Beyond the {Nav-Graph}: Vision-and-language navigation in continuous
  environments.
\newblock In \emph{ECCV}, 2020.

\bibitem[Lee et~al.(2024)Lee, Miyanishi, Kurita, Sakamoto, Azuma, Matsuo, and
  Inoue]{lee2024citynav}
Jungdae Lee, Taiki Miyanishi, Shuhei Kurita, Koya Sakamoto, Daichi Azuma,
  Yutaka Matsuo, and Nakamasa Inoue.
\newblock {CityNav}: Language-goal aerial navigation dataset with geographic
  information.
\newblock \emph{arXiv preprint arXiv:2406.14240v1}, 2024.

\bibitem[Lee et~al.(2025)Lee, Miyanishi, Kurita, Sakamoto, Azuma, Matsuo, and
  Inoue]{lee2025citynav}
Jungdae Lee, Taiki Miyanishi, Shuhei Kurita, Koya Sakamoto, Daichi Azuma,
  Yutaka Matsuo, and Nakamasa Inoue.
\newblock {CityNav}: A large-scale dataset for real-world aerial navigation.
\newblock In \emph{ICCV}, 2025.

\bibitem[Lei et~al.(2025)Lei, Yang, Chen, Li, and Liu]{lei2024scaffold}
Xuanyu Lei, Zonghan Yang, Xinrui Chen, Peng Li, and Yang Liu.
\newblock Scaffolding coordinates to promote vision-language coordination in
  large multi-modal models.
\newblock In \emph{COLING}, 2025.

\bibitem[Li et~al.(2026)Li, Dong, Ma, Chen, Zhao, and Zhou]{li2026closer}
Shaoxuan Li, Xiangyu Dong, Xiaoguang Ma, Junfeng Chen, Haoran Zhao, and Yaoming
  Zhou.
\newblock {CLOSER-VLN}: Closed-loop self-verified retrieval-augmented reasoning
  for aerial vision-language navigation.
\newblock \emph{arXiv preprint arXiv:2606.28397}, 2026.

\bibitem[Lin et~al.(2025)Lin, Wei, An, Gao, Zou, Luo, Huang, Zhang, and
  Li]{lin2025drawunderstand}
Weifeng Lin, Xinyu Wei, Ruichuan An, Peng Gao, Bocheng Zou, Yulin Luo, Siyuan
  Huang, Shanghang Zhang, and Hongsheng Li.
\newblock Draw-and-understand: Leveraging visual prompts to enable {MLLMs} to
  comprehend what you want.
\newblock In \emph{ICLR}, 2025.

\bibitem[Liu et~al.(2023)Liu, Zhang, Qi, Wang, Zhang, and Wu]{liu2023aerialvln}
Shubo Liu, Hongsheng Zhang, Yuankai Qi, Peng Wang, Yanning Zhang, and Qi Wu.
\newblock {AerialVLN}: Vision-and-language navigation for {UAVs}.
\newblock In \emph{ICCV}, 2023.

\bibitem[Liu et~al.(2024)Liu, Zeng, Ren, Li, Zhang, Yang, Jiang, Li, Yang, Su,
  Zhu, and Zhang]{liu2024groundingdino}
Shilong Liu, Zhaoyang Zeng, Tianhe Ren, Feng Li, Hao Zhang, Jie Yang, Qing
  Jiang, Chunyuan Li, Jianwei Yang, Hang Su, Jun Zhu, and Lei Zhang.
\newblock Grounding {DINO}: Marrying {DINO} with grounded pre-training for
  open-set object detection.
\newblock In \emph{ECCV}, 2024.

\bibitem[Miyanishi et~al.(2023)Miyanishi, Kitamori, Kurita, Lee, Kawanabe, and
  Inoue]{miyanishi2023cityrefer}
Taiki Miyanishi, Fumiya Kitamori, Shuhei Kurita, Jungdae Lee, Motoaki Kawanabe,
  and Nakamasa Inoue.
\newblock {CityRefer}: Geography-aware {3D} visual grounding dataset on
  city-scale point cloud data.
\newblock In \emph{NeurIPS Datasets and Benchmarks Track}, 2023.

\bibitem[Nasiriany et~al.(2024)Nasiriany, Xia, Yu, Xiao, Liang, Dasgupta, Xie,
  Driess, Wahid, Xu, et~al.]{nasiriany2024pivot}
Soroush Nasiriany, Fei Xia, Wenhao Yu, Ted Xiao, Jacky Liang, Ishita Dasgupta,
  Annie Xie, Danny Driess, Ayzaan Wahid, Zhuo Xu, et~al.
\newblock {PIVOT}: Iterative visual prompting elicits actionable knowledge for
  {VLMs}.
\newblock In \emph{ICML}, 2024.

\bibitem[Sathyamoorthy et~al.(2024)Sathyamoorthy, Weerakoon, Elnoor, Zore,
  Ichter, Xia, Tan, Yu, and Manocha]{sathyamoorthy2024convoi}
Adarsh~Jagan Sathyamoorthy, Kasun Weerakoon, Mohamed Elnoor, Anuj Zore, Brian
  Ichter, Fei Xia, Jie Tan, Wenhao Yu, and Dinesh Manocha.
\newblock {CoNVOI}: Context-aware navigation using vision language models in
  outdoor and indoor environments.
\newblock In \emph{IROS}, 2024.

\bibitem[Shtedritski et~al.(2023)Shtedritski, Rupprecht, and
  Vedaldi]{shtedritski2023redcircle}
Aleksandar Shtedritski, Christian Rupprecht, and Andrea Vedaldi.
\newblock What does {CLIP} know about a red circle? {V}isual prompt engineering
  for {VLMs}.
\newblock In \emph{ICCV}, 2023.

\bibitem[Sun et~al.(2026)Sun, Liu, Su, Zhu, Gu, Zou, Liu, Xia, Du, and
  Xu]{sun2025refdrone}
Zhichao Sun, Yepeng Liu, Zhiling Su, Huachao Zhu, Yuliang Gu, Yuda Zou, Zelong
  Liu, Gui-Song Xia, Bo Du, and Yongchao Xu.
\newblock {RefDrone}: A challenging benchmark for referring expression
  comprehension in drone scenes.
\newblock In \emph{CVPR Findings}, 2026.

\bibitem[Tong et~al.(2026)Tong, Dong, Ma, Zhao, Zhou, and Lin]{tong2026visa}
Haoyu Tong, Xiangyu Dong, Xiaoguang Ma, Haoran Zhao, Yaoming Zhou, and Chenghao
  Lin.
\newblock {ViSA}-enhanced aerial {VLN}: A visual-spatial reasoning enhanced
  framework for aerial vision-language navigation.
\newblock \emph{arXiv preprint arXiv:2603.08007}, 2026.

\bibitem[Wang et~al.(2026{\natexlab{a}})Wang, Li, Zhang, He, Jiang, Li, Zhao,
  Fan, Hou, Liang, Wen, and Gu]{wang2026consistnav}
Haosen Wang, Zhenyang Li, Yinqiang Zhang, Zongqi He, Lutao Jiang, Kai Li,
  Yizhou Zhao, Liaoyuan Fan, Wenjian Hou, Tingbang Liang, Yibin Wen, and Defeng
  Gu.
\newblock {ConsistNav}: Closing the action consistency gap in zero-shot object
  navigation with semantic executive control.
\newblock \emph{arXiv preprint arXiv:2605.09869}, 2026{\natexlab{a}}.

\bibitem[Wang et~al.(2026{\natexlab{b}})Wang, Sun, He, and Zhu]{wang2026raco}
Sen Wang, Yiming Sun, Jiaxuan He, and Pengfei Zhu.
\newblock {RACO}: Reliability-aware coarse-goal optimization for
  inspection-oriented {UAV} vision-language navigation.
\newblock \emph{arXiv preprint arXiv:2608.22678}, 2026{\natexlab{b}}.

\bibitem[Wu et~al.(2024)Wu, Zhang, Xia, Li, Xia, Chang, Yu, Kim, Rossi, Zhang,
  Mitra, Metaxas, Yao, Shang, and McAuley]{wu2024vpsurvey}
Junda Wu, Zhehao Zhang, Yu Xia, Xintong Li, Zhaoyang Xia, Aaron Chang, Tong Yu,
  Sungchul Kim, Ryan~A. Rossi, Ruiyi Zhang, Subrata Mitra, Dimitris~N. Metaxas,
  Lina Yao, Jingbo Shang, and Julian McAuley.
\newblock Visual prompting in multimodal large language models: A survey.
\newblock \emph{arXiv preprint arXiv:2409.15310}, 2024.

\bibitem[Xu et~al.(2026{\natexlab{a}})Xu, Hu, Gao, Zhu, Zhao, and
  Yin]{xu2026geonav}
Haotian Xu, Yue Hu, Chen Gao, Zhengqiu Zhu, Yong Zhao, and Quanjun Yin.
\newblock {GeoNav}: Empowering {MLLMs} with dual-scale geospatial reasoning for
  language-goal aerial navigation.
\newblock \emph{Pattern Recognition}, 177:\penalty0 113365, 2026{\natexlab{a}}.

\bibitem[Xu et~al.(2026{\natexlab{b}})Xu, Hu, Zhu, Zhao, Wang, and
  Yin]{xu2026sbfnav}
Haotian Xu, Yue Hu, Zhengqiu Zhu, Yong Zhao, Wenguan Wang, and Quanjun Yin.
\newblock Map the possibilities: Spatial belief fields for language-goal aerial
  navigation.
\newblock \emph{arXiv preprint arXiv:2609.05841}, 2026{\natexlab{b}}.

\bibitem[Xue et~al.(2026)Xue, Yu, Shen, Hu, Wang, Yang, Wang, and
  Sun]{xue2026seereach}
Fanfu Xue, En Yu, Yantian Shen, Zhikun Hu, Hongjun Wang, Yang Yang, Xindi Wang,
  and Jiande Sun.
\newblock See-and-reach: Precise vision-language navigation for {UAVs} within
  the field of view.
\newblock \emph{arXiv preprint arXiv:2606.20045}, 2026.

\bibitem[Yan et~al.(2024)Yan, Yang, Wu, Zhu, Yang, Li, Lin, Wang, McAuley, Gao,
  and Wang]{yan2024somllava}
An Yan, Zhengyuan Yang, Junda Wu, Wanrong Zhu, Jianwei Yang, Linjie Li, Kevin
  Lin, Jianfeng Wang, Julian McAuley, Jianfeng Gao, and Lijuan Wang.
\newblock List items one by one: A new data source and learning paradigm for
  multimodal {LLMs}.
\newblock In \emph{COLM}, 2024.

\bibitem[Yang et~al.(2023)Yang, Zhang, Li, Zou, Li, and Gao]{yang2023som}
Jianwei Yang, Hao Zhang, Feng Li, Xueyan Zou, Chunyuan Li, and Jianfeng Gao.
\newblock Set-of-mark prompting unleashes extraordinary visual grounding in
  {GPT-4V}.
\newblock \emph{arXiv preprint arXiv:2310.11441}, 2023.

\bibitem[Zhang et~al.(2025{\natexlab{a}})Zhang, Khayatkhoei, Chhikara, and
  Ilievski]{zhang2025mllmsknow}
Jiarui Zhang, Mahyar Khayatkhoei, Prateek Chhikara, and Filip Ilievski.
\newblock {MLLMs} know where to look: Training-free perception of small visual
  details with multimodal {LLMs}.
\newblock In \emph{ICLR}, 2025{\natexlab{a}}.

\bibitem[Zhang et~al.(2025{\natexlab{b}})Zhang, Cai, Zhang, Zhuang, Li, and
  Mao]{zhang2024earthmarker}
Wei Zhang, Miaoxin Cai, Tong Zhang, Yin Zhuang, Jun Li, and Xuerui Mao.
\newblock {EarthMarker}: A visual prompting multi-modal large language model
  for remote sensing.
\newblock \emph{IEEE Transactions on Geoscience and Remote Sensing}, 63,
  2025{\natexlab{b}}.

\bibitem[Zhang et~al.(2025{\natexlab{c}})Zhang, Yadav, Han, and
  Shutova]{zhang2025crossmodal}
Zhi Zhang, Srishti Yadav, Fengze Han, and Ekaterina Shutova.
\newblock Cross-modal information flow in multimodal large language models.
\newblock In \emph{CVPR}, 2025{\natexlab{c}}.

\end{thebibliography}
}

\clearpage
\ifdefined\arxivversion\else\setcounter{page}{1}\fi%
\maketitlesupplementary
\setcounter{section}{0}
\renewcommand{\thesection}{\Alph{section}}%

\section{Implementation details}
\label{sec:impl}

\paragraph{Text input.}
Besides the image with the drawn landmarks, the text input contains a short description of the view (altitude, coverage, orientation, and the meaning of the yellow marks), the original instruction, and the parsed target, landmarks, and surroundings of the revised annotations.
Removing the parsed fields changes SR by less than 0.3 points.

\paragraph{Training.}
We use LoRA with rank 16, $\alpha{=}32$, and dropout 0.05 on all linear layers of the language model.
Images are resized to 448\,px.
We train for one epoch on 60k samples with AdamW, a learning rate of $10^{-4}$, 50 warm-up steps followed by linear decay, and a total batch size of 8 on two RTX 5880 Ada GPUs.
The loss is computed on the answer tokens only.
At test time we decode greedily.

\paragraph{Evaluation.}
CityNav evaluation is deterministic: running HETT twice gives identical results for every episode.
In \cref{tab:marks}, the distractor box is the highest-scoring GroundingDINO detection that lies more than 15\,m from the target.

\paragraph{Attention analysis.}
The 1,000 end points are drawn with a fixed seed from the test end points of HETT whose target is inside the view.
The similar objects are the GroundingDINO detections of the target phrase that do not overlap the target, lie more than 10\,m from it, and are smaller than 50\,m (at most ten per view).
We compare objects by attention per unit area, since buildings are much larger than cars.
For the AUC, each view is scored by the attention per unit area on the target box, divided by the same quantity when the image is paired with the generic prompt ``Describe the image.''; this discounts the salience of the drawn marks.

\section{Additional results}
\label{sec:addres}
This section gives the exact numbers behind \cref{fig:gap,fig:interaction} and the comparison of last-step selectors in \cref{sec:selector}. All navigator results are from our runs of the released code and weights.

\paragraph{Revised annotations.}
The revision by HETT~\cite{ding2026hett} changes the landmark lists and removes 30 of the 5,311 test-unseen episodes; descriptions, trajectories, and targets are identical.
Because the revision raises the training-free landmark-centroid rule much more on test-unseen (+9.7 points on the 5,281 shared episodes) than on validation (+2 points), we checked whether it uses information about the target.
Replaying the edits one type at a time, about 8.9 of the 9.7 points come from fixing extraction errors: empty landmark lists (411 of the 5,281 kept test episodes; 26 of the 30 removed episodes also have no matched landmark), misspelled names that were matched to distant map entities, and landmarks that the description mentions but the original list missed.
At most 0.8 points come from edits that could reflect knowledge of the target, such as adding a landmark that the description does not mention.
We therefore use the revised annotations for the main results and report the original ones alongside.

\begin{table}[t]
\centering
\small
\setlength{\tabcolsep}{3pt}
\begin{tabular}{lccccc}
\toprule
 & & \multicolumn{2}{c}{Among failures} & Reached & Recovered \\
\cmidrule(lr){3-4}
Navigator & SR & $\le$50\,m & in view & but failed & (in view) \\
\midrule
Seq2Seq & 1.2 & 4.0 & 5.3 & -- & 53.7 \\
CMA & 1.0 & 5.5 & 7.1 & -- & 50.4 \\
MGP$^*$ & 13.7 & 48.0 & 55.9 & -- & 61.7 \\
FlightGPT & 24.2 & 51.1 & 56.2 & 17.3 & 57.6 \\
HETT & 25.9 & 59.7 & 66.5 & 21.5 & 59.5 \\
LM-centroid & 40.2 & 73.7 & 78.8 & -- & 57.5 \\
\bottomrule
\end{tabular}
\caption{\textbf{Where navigators fail} (test-unseen, \%).
``$\le$50\,m'' and ``in view'': share of failed episodes that end within 50\,m of the target, or with the target inside the drone's final view from 50\,m.
``Reached but failed'': share of all episodes that pass within 20\,m of the target but stop outside; this needs per-step trajectories, which we have for FlightGPT and HETT.
``Recovered'': share of the in-view failures that our last-step module turns into successes. $^*$Where the drone stops, before CityNav's goal-selection step (\cref{sec:mgp}).}
\label{tab:decomp}
\end{table}

\paragraph{Target swap.}
\label{sec:swap}
In \cref{fig:gap}b we replace the first mention of the target in the description with the target phrase of an object of another category on the same map, which changes 5,199 test-unseen descriptions.
Since the landmarks and the spatial relations are kept, a navigator that uses the target should fly to an object of the new category next to the same landmarks.
We call an object \emph{next to the landmarks} if its distance to the nearest named landmark is at most that of the original target, or 20\,m if this is smaller.
For 1,099 episodes, such an object exists and the nearest one lies more than 20\,m from the original target; in the other episodes, the swapped instruction either describes an object that is already within 20\,m or one that does not exist next to the landmarks.
We designate the nearest such object as the new target and judge success as in CityNav, by a final position within 20\,m of its center.
In 386 of the 1,099 episodes there is exactly one candidate, so the new target is unambiguous; \cref{fig:gap}b and \cref{tab:perturb} use these episodes.
The replaced phrase can make a description less fluent (\eg ``Two story white car with a gray roof''), but all methods receive the same descriptions.
\Cref{tab:perturb} also reports two other subsets, and the conclusion does not change.
Matching the color and other attribute words of the new phrase against each candidate's own descriptions makes 152 more episodes unambiguous; on all 538 episodes, the gain is +17.1 [+13.4, +20.8] for \ours and between $-0.9$ and $+0.0$ for every navigator.
As a control, replacing the landmark names instead, with those of another description on the same map, moves HETT's end point by a median of 164\,m and lowers its SR from 25.9\% to 2.1\% (5,047 episodes).

\begin{table}[t]
\centering
\small
\setlength{\tabcolsep}{1.8pt}
\begin{tabular}{lcccc}
\toprule
 & \makecell{Median\\shift (m)} & \makecell{Gain in new-\\target SR (386)} & \makecell{Gain\\(1,099)} & \makecell{Gain, $>$40\,m\\(688)} \\
\midrule
Seq2Seq$^\ddagger$ & 0.0 & +0.0 [+0.0, +0.0] & +0.0 & +0.0 \\
CMA$^\ddagger$ & 0.0 & +0.0 [+0.0, +0.0] & +0.1 & +0.0 \\
MGP & 0.0 & $-$1.3 [$-$2.8, +0.0] & $-$0.5 & $-$0.3 \\
HETT & 3.9 & +0.0 [$-$2.3, +2.3] & $-$0.3 & $-$1.6 \\
FlightGPT & 17.7 & $-$0.5 [$-$4.4, +3.4] & $-$0.4 & +0.4 \\
\midrule
\ours & 22.1 & \textbf{+20.5} [+15.8, +25.1] & \textbf{+13.1} & \textbf{+13.1} \\
\bottomrule
\end{tabular}
\caption{\textbf{Target swap} (test-unseen). Main columns: the 386 episodes with exactly one candidate new target (\cref{sec:swap}). Shift: distance between the end points with the original and the swapped instruction. With the original instruction, a stop can be near the new object by chance. Gain: SR for the new target with the swapped minus with the original instruction (10.1 $\rightarrow$ 30.6 for \ours, applied at HETT's end points), with paired bootstrap 95\% confidence intervals. Last two columns: all 1,099 episodes with a designated new target, and those whose new and original targets are more than 40\,m apart, so that the two success areas do not overlap. $^\ddagger$371, 1,058, and 658 episodes kept by the official VLN-CE loader.}
\label{tab:perturb}
\end{table}

\begin{table}[t]
\centering
\small
\setlength{\tabcolsep}{5pt}
\resizebox{\linewidth}{!}{%
\begin{tabular}{lccc}
\toprule
 & \multicolumn{2}{c}{Test} & Val \\
\cmidrule(lr){2-3}\cmidrule(lr){4-4}
Last step at HETT's end point & SR & on LM & SR \\
\midrule
Stop (HETT) & 25.94 & -- & 23.40 \\
GroundingDINO, target phrase & 27.84 & -- & 23.14 \\
\midrule
Zero-shot VLM, no landmarks drawn & 36.91 & 17.4 & 33.45 \\
Zero-shot VLM, landmarks drawn & 39.06 & 23.1 & 35.05 \\
Fine-tuned, no landmarks drawn & 43.53 & 14.9 & 41.89 \\
Fine-tuned, landmarks drawn (\ours) & \textbf{51.30} & 16.3 & \textbf{47.65} \\
\bottomrule
\end{tabular}}
\caption{\textbf{Anchors $\times$ recognition supervision} (SR, \%). ``on LM'': share of predicted boxes whose center falls inside a drawn landmark outline. Interaction on test: $(51.30-43.53)-(39.06-36.91)=+5.6$ [+4.0, +7.2]; on val: $(47.65-41.89)-(35.05-33.45)=+4.2$ [+2.5, +5.7]. All fine-tuned models use the same 60k samples, and all models the same prompt.}
\label{tab:ladder}
\end{table}

\begin{table}[t]
\centering
\small
\begin{tabular}{lcc}
\toprule
Selector at MGP's predicted goal & Revised & Original \\
\midrule
None (stop at the goal) & 21.95 & 17.17 \\
GroundingDINO, as released~\cite{lee2024citynav,liu2024groundingdino} & \phantom{0}8.46 & \phantom{0}6.57 \\
GroundingDINO, coordinates fixed & 22.44 & 18.19 \\
\ours & 49.02 & 38.13 \\
\ours + LA & \textbf{52.77} & \textbf{42.14} \\
\bottomrule
\end{tabular}
\caption{\textbf{Last-step selectors at the same viewpoints} (test-unseen SR, \%). Revised: 5,281 episodes; original: all 5,311 episodes. The two GroundingDINO rows differ only in the conversion from the box to world coordinates (\cref{sec:mgp}). The comparison is slightly unfavorable to \ours, because we charge the flight from the drone's actual position to the predicted goal, which the official step does not.}
\label{tab:selector}
\end{table}

\section{Alternatives that do not help}
\label{sec:nothelp}
Several natural alternatives do not improve over a single prediction at the end point:
\begin{itemize}
\item \textbf{Numbered candidates.} Letting the VLM pick one of about ten detector proposals, in the style of Set-of-Mark, is no better than predicting a box; even when the true target is among them, it is picked only 34--41\% of the time.
\item \textbf{Coarse-to-fine or two views.} Re-asking on a crop around the first answer does not fix a wrong first choice, and showing a wide and a close-up view together makes the coordinates drift.
\item \textbf{Stopping during flight.} Stopping when the model first detects the target is 15 to 18 points worse than deciding once at the end, because the model is over-confident early.
\end{itemize}
Descending to look closer does not help either, since a lower view is only a smaller crop of the same orthophoto; this is why \ours climbs instead.

\section{Failure analysis}
\label{sec:fail}
\begin{figure*}[t]
\centering
\includegraphics[width=\textwidth]{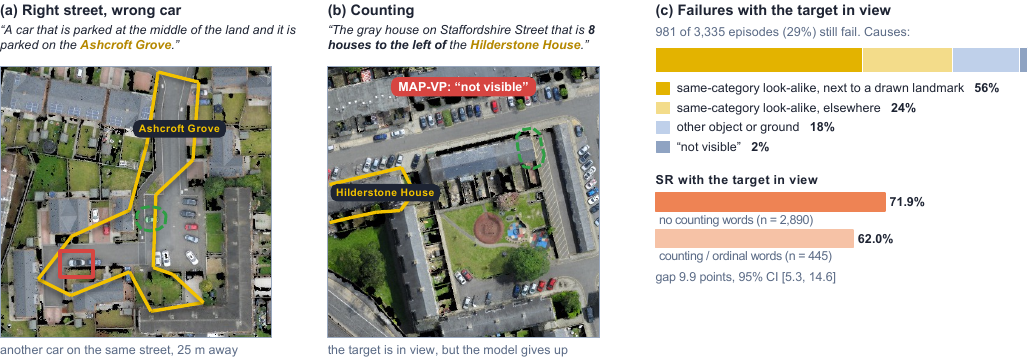}
\caption{\textbf{Failures of \ours when the target is in view} (test-unseen, HETT end points, view from 50\,m). Dashed green: the target; red: the choice of \ours; yellow: the drawn landmarks. (a)~A typical failure: the chosen car is on the named street but is not the target. (b)~A description that requires counting. (c)~Causes of all such failures, and SR with and without counting or ordinal words in the description.}
\label{fig:fail}
\end{figure*}
We analyze the 3,335 test episodes in which the target lies entirely inside HETT's final view from 50\,m.
\ours still fails in 981 of them (29\%), and \cref{fig:fail} groups these failures by cause.
In 80\% of them, the chosen box falls on another object of the same category.
In most of these cases (56\% of all failures), the chosen object lies on or next to a drawn landmark, as in \cref{fig:fail}a.
The model has thus found the right place but not the right instance: a street or a parking lot often contains several objects that match the description.
Another 18\% of the boxes fall on objects of other categories or on the ground, and only 2\% of the failures are wrong ``not visible'' answers.
Descriptions that count objects, such as ``8 houses to the left of the Hilderstone House'' (\cref{fig:fail}b), are harder.
They appear in 13\% of the in-view episodes, and SR on them is 62.0\% instead of 71.9\% (gap 9.9 points, 95\% CI [5.3, 14.6]).
Telling instances apart along a landmark, for example by counting, is therefore the main remaining weakness of the last step.

\section{Reproducing MGP}
\label{sec:mgp}

Running the released MGP code and checkpoint~\cite{lee2024citynav} as is does not reproduce the reported results.
We traced the gap to two issues in the released code.
This section explains them, shows that fixing the first one reproduces the reported numbers (\cref{tab:mgp}), and states which setting each of our results uses.

\paragraph{How MGP produces its final position.}
At each step, MGP predicts a goal on the map, and the drone executes five primitive actions towards it (turn by 30$^\circ$ or move forward 5\,m).
After the last step, CityNav's goal-selection step detects the target phrase with GroundingDINO in a 50\,m top-down view centered at the last predicted goal, and the drone is placed at the center of the most confident box.
This placement is not simulated as flight and is not charged to SPL.
If nothing is detected, the whole view is used as the box.

\paragraph{Issue 1: the drone turns in place.}
In the released \texttt{move()} function, all five actions are chosen from the pose at the beginning of the step, not from the pose after the previous action.
When the goal is more than 30$^\circ$ off the heading, the drone therefore turns five times in a row, by 150$^\circ$ in total, and overshoots.
On test-unseen, the median flight path is only 50\,m, and the drone stops a median of 92\,m from its own predicted goal.
The same function in the HETT code base~\cite{ding2026hett} chooses each action from the updated pose; we use this version.
The median path then becomes 190\,m, and the drone stops a median of 17\,m from its predicted goal.
After this fix, the oracle success rate matches the reported one (26.00 \vs 26.04, \cref{tab:mgp}), so the reported results were obtained without this issue.

\paragraph{Issue 2: the selected box is mapped to the wrong place.}
The function that converts the selected box from pixels to world coordinates (\texttt{xyxy\_to\_global\_bbox}) has three errors: it starts from the center of the view instead of its corner, it uses half of the true meters per pixel, and it swaps rows and columns.
The final position therefore lies 49\,m on average from the box that GroundingDINO selected, and 35\,m from the view center when nothing is detected.
Keeping this issue reproduces the reported NE and SR (\cref{tab:mgp}), so the reported results include it.

\begin{table}[h]
\centering
\small
\setlength{\tabcolsep}{4pt}
\begin{tabular}{lccc}
\toprule
Setting (original annotations) & NE (m) & SR & OSR \\
\midrule
Reported~\cite{lee2024citynav} & 93.8 & 6.38 & 26.04 \\
Released code as is & 87.5 & 5.91 & 11.11 \\
Issue 1 fixed (\textbf{our setting}) & 94.1 & 6.57 & 26.00 \\
Issues 1 and 2 fixed & 85.4 & 18.19 & 32.48 \\
\bottomrule
\end{tabular}
\caption{\textbf{Reproducing MGP} with the released checkpoint (all 5,311 episodes). The final position is the output of CityNav's goal-selection step, as in the official evaluation. OSR counts an episode if the trajectory or the final position comes within 20\,m of the target.}
\label{tab:mgp}
\end{table}

\paragraph{Which setting each result uses.}
\begin{itemize}
\item \textbf{MGP's own results} (\cref{tab:main,tab:sota}) use the setting that reproduces the reported numbers: issue~1 fixed, the official goal-selection step unchanged, including issue~2. This keeps our MGP numbers comparable with other papers. With the revised annotations, this setting gives 8.46\% on test-unseen, close to the 9.70\% reported by HTNav~\cite{htnav2026}.
\item \textbf{MGP + \ours} (\cref{tab:main,tab:sota}) replaces the goal-selection step: \ours starts where the drone stops, or takes over its last steps when the step limit requires it (\cref{sec:budget}), and its flight is charged to SPL. Where the drone stops, without any goal selection, SR is already 13.69\% (revised); the remaining 27.9 points of the gain to 41.56\% come from \ours itself.
\item \textbf{Failure analysis} (\cref{fig:gap,fig:distance}, \cref{tab:decomp}) also measures MGP where the drone stops, since this is the view that \ours receives.
\item \textbf{Same-viewpoint comparison} (\cref{tab:selector}) runs every selector at MGP's last predicted goal, which is where the official goal-selection step looks. Here we report GroundingDINO both as released and with issue~2 fixed. The fixed version is the fair baseline: as released, GroundingDINO lowers SR from 21.95\% to 8.46\%, which would favor \ours unfairly.
\end{itemize}

\section{Information used by each method}
\label{sec:fair}

\begin{figure*}[t]
\centering
\includegraphics[width=\textwidth]{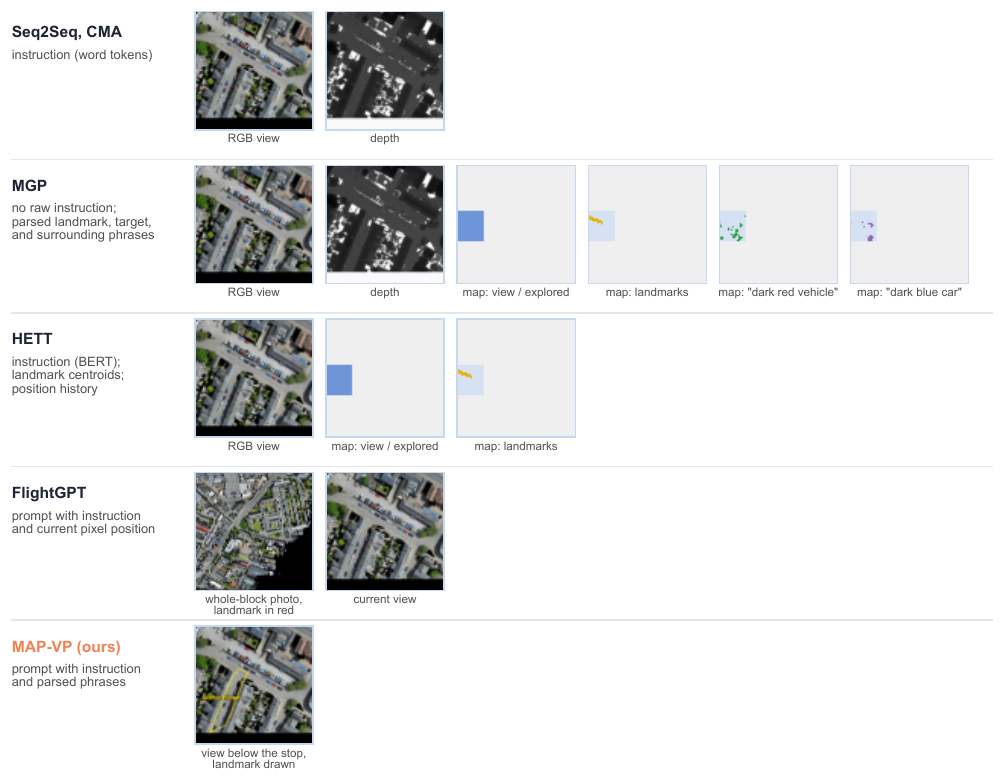}
\caption{\textbf{Every input of each method} for the test episode of \cref{fig:teaser} (``A dark red vehicle parked along Grafton Street across from a dark blue car''). All inputs are produced by each method's released code at the same position, HETT's end point, 50\,m above the ground. Maps cover the whole block (410\,m). Every navigator sees the same ground as \ours; \ours additionally draws the landmark that the navigators already receive as a map, a centroid, or a painted mask.}
\label{fig:fair}
\end{figure*}

\ours adds no information that the navigators it is applied to do not already receive.
\Cref{tab:fair} lists every input of each method, read from the released code that we run (\cref{sec:setting}), and \cref{fig:fair} shows all image and map inputs for one test episode, each produced by the method's own code.

\paragraph{Shared setting.}
CityNav renders every view from a geo-referenced orthophoto at the drone's pose, so pixels and world coordinates are related by a fixed transform for all methods.
CityNav also provides, for each instruction, the names of the mentioned landmarks and their OpenStreetMap polygons; all methods that use them read the same lists.

\paragraph{Same information, smaller area.}
MGP, HETT, and FlightGPT receive the landmark polygons both in training and at test time, and over the whole block: MGP and HETT as a rasterized map of 410\,m$\times$410\,m, HETT also as landmark centroids, and FlightGPT as red masks on an orthophoto of the whole block.
\ours draws the same polygons, but only into the 100\,m$\times$100\,m view below the drone's final position, plus one view from 100\,m when it looks again.
This view shows the same ground that every navigator sees at that position (\cref{fig:fair}).
For these navigators, \ours therefore adds no information.
Seq2Seq and CMA use neither the polygons nor the pose; for them, \ours also adds the polygons.

\paragraph{What \ours does not use.}
At test time, \ours never uses the target's position, never views places that the drone has not reached, and calls no external service: it runs the open Qwen3-VL-4B model.
Like the navigators, it is trained only on the training split (\cref{sec:data}).
All flight that \ours adds, to the chosen box and for looking again, is charged to SPL.

\begin{table*}[t]
\centering
\footnotesize
\setlength{\tabcolsep}{3pt}
\renewcommand{\arraystretch}{1.15}
\begin{tabular}{>{\raggedright\arraybackslash}p{1.7cm}>{\raggedright\arraybackslash}p{2.4cm}>{\raggedright\arraybackslash}p{2.7cm}>{\raggedright\arraybackslash}p{3.2cm}>{\raggedright\arraybackslash}p{2.6cm}>{\centering\arraybackslash}p{1.3cm}>{\raggedright\arraybackslash}p{1.5cm}}
\toprule
Method & Text & Images (every step) & Maps and coordinates (every step) & Form of the landmark input & \makecell{Landmark\\polygons\\(train / test)} & Output \\
\midrule
Seq2Seq, CMA & instruction (word tokens) & current RGB and depth & -- & -- & -- / -- & actions \\
MGP & parsed landmark, target, and surrounding phrases (no raw instruction) & current RGB and depth & whole-block maps: view, explored area, landmarks, detections of the target and surrounding phrases & polygons filled into a map channel & \checkmark{} / \checkmark{} & goal, then actions \\
HETT & instruction (BERT) & current RGB & whole-block maps: view, explored area, landmarks; landmark centroids; position history & polygons filled into a map channel; centroid coordinates as numbers & \checkmark{} / \checkmark{} & goal and actions \\
FlightGPT & instruction and current pixel position & whole-block orthophoto with landmarks painted; current view & -- & polygons filled in red and names written on the orthophoto & \checkmark{} / \checkmark{} & goal \\
LM-centroid & -- & -- & landmark polygons & mean of the polygon vertices, used as the goal & -- / \checkmark{} & goal \\
\midrule
\ours & instruction and parsed phrases & view below the final position, with landmarks drawn & -- & outlines and names drawn on the current view & \checkmark{} / \checkmark{} & box, then one move \\
\bottomrule
\end{tabular}
\caption{\textbf{Inputs of each method}, read from the released code; training and test inputs are the same. All views are rendered from the same geo-referenced orthophoto at the drone's pose, and all methods that use landmarks read the same lists of landmark names and polygons provided by CityNav. No method receives the polygon coordinates as text. When \ours looks again, it also receives the view from 100\,m.}
\label{tab:fair}
\end{table*}

\section{Step budget}
\label{sec:budget}

The released code of each navigator ends an episode after a fixed number of steps, and \ours runs after the navigator stops.
To keep the comparison fair, \ours stays within this step limit in all our results (\cref{sec:act}).
This section gives the details and compares the results with a flight that is not counted in steps (\cref{tab:budget}).

\paragraph{Protocol.}
\ours flies to its predicted point with the navigator's own primitive actions (move forward 5\,m or turn by 30$^\circ$, chosen by the same look-ahead controller) and may use 20 of them.
Looking again adds one step for the climb.
If the navigator has fewer actions left, \ours takes over earlier: it starts from the navigator's position at the last step that leaves this budget, and the remaining steps of the navigator are not executed.
For HETT, \ours thus replaces the last two steps (three with look-again).
MGP executes five actions per step, so \ours replaces its last four steps (five).
Seq2Seq and CMA end near the ground and climb in steps of 2\,m, so their budget also covers the climb back to 50\,m, and looking again costs 25 actions.
FlightGPT always runs two predictions of at most 75 actions each and never uses all 150, so \ours only uses the actions that are left.
Its heading at the end is not recorded, and we assume the worst case, facing away from the predicted point.
LM-centroid has the same limit as HETT and always ends with at least 90 actions left.
Whenever \ours starts from the same position as the navigator's end point, the view and the answer of the model are the same.
The validation results in \cref{tab:main} and the original-annotation results in \cref{tab:sota} follow the same protocol.

\paragraph{Results.}
\Cref{tab:budget} compares our results with a version in which the drone flies straight to the predicted point and this flight is not counted in steps.
The two differ by less than one point in SR for every navigator.
HETT reaches its limit in almost every episode, but its last steps barely move the drone (median 0\,m, as it mostly turns in place), so taking them over costs nothing.
A smaller budget of 10 actions, \ie one step of HETT, does not reach the farther boxes and lowers SR to 49.20 for HETT and to 40.60 for MGP.

\begin{table*}[t]
\centering
\footnotesize
\setlength{\tabcolsep}{4pt}
\renewcommand{\arraystretch}{1.15}
\begin{tabular}{lllcccccc}
\toprule
& & & \multicolumn{2}{c}{Earlier start (\%)} & \multicolumn{2}{c}{+\,\ours} & \multicolumn{2}{c}{+\,\ours\,+\,LA} \\
\cmidrule(lr){4-5}\cmidrule(lr){6-7}\cmidrule(lr){8-9}
Navigator & Step limit (released code) & Taken over by \ours & w/o LA & w/ LA & \cref{tab:main} & Not counted & \cref{tab:main} & Not counted \\
\midrule
Seq2Seq & 200 actions & climb + 20 actions & 52.0 & 61.1 & 3.87 / 2.28 & 3.89 / 2.22 & 6.98 / 3.65 & 6.98 / 3.43 \\
CMA & 200 actions & climb + 20 actions & 41.1 & 48.9 & 4.61 / 2.18 & 4.71 / 2.20 & 9.19 / 3.85 & 10.03 / 3.99 \\
MGP & 20 steps $\times$ 5 actions & last 4 steps (5 with LA) & 34.2 & 34.9 & 41.56 / 34.72 & 42.06 / 34.85 & 48.31 / 38.87 & 49.06 / 39.07 \\
FlightGPT & 2 steps $\times$ 75 actions & actions left only & 0 & 0 & 45.05 / 36.51 & 44.84 / 35.85 & 48.85 / 38.18 & 48.70 / 37.52 \\
HETT & 20 steps $\times$ 10 actions & last 2 steps (3 with LA) & 98.7 & 98.8 & 51.56 / 38.31 & 51.30 / 37.51 & 54.55 / 39.96 & 53.95 / 38.59 \\
LM-centroid & 20 steps $\times$ 10 actions & actions left only & 0 & 0 & 59.55 / 52.64 & 59.19 / 51.74 & 61.37 / 53.49 & 61.05 / 52.60 \\
\bottomrule
\end{tabular}
\caption{\textbf{Step budget} (test-unseen, SR / SPL). \cref{tab:main}: our results, in which the drone flies with the navigator's own actions, at most 20 actions plus one step to look again, and the whole episode stays within the step limit of the navigator's released code. Not counted: the drone flies straight to the predicted point, and this flight is added to the path length but not counted in steps. Earlier start: share of episodes in which \ours takes over before the navigator ends.}
\label{tab:budget}
\end{table*}

\section{Results in the standard CityNav format}
\label{sec:full}

Prior work on CityNav~\cite{lee2025citynav,ding2026hett,htnav2026,xu2026sbfnav} reports NE, SR, OSR, and SPL on the three evaluation splits, and several methods also report the three difficulty levels of test-unseen that CityNav defines by the distance from the start to the target.
\Cref{tab:full,tab:difficulty} give our results in this format, with \ours applied as in \cref{tab:main} (look-again included, within each navigator's step limit).
NE is the mean 2D distance between the final position and the target.
OSR counts an episode if the drone passes within 20\,m of the target at any point, including the flight of \ours.
The validation numbers use the held-out episodes (2,320 in val-seen and 2,547 in val-unseen), since 300 other validation episodes were used to choose the look-again rule.
MGP uses CityNav's official goal-selection step, as in \cref{tab:main}.
\ours improves every metric of every navigator on every split and on every difficulty level; for HETT, SR rises by 26.4, 27.8, and 32.5 points on the easy, medium, and hard levels.

\begin{table*}[t]
\centering
\footnotesize
\setlength{\tabcolsep}{3.5pt}
\renewcommand{\arraystretch}{1.1}
\begin{tabular}{lrrrrrrrrrrrr}
\toprule
& \multicolumn{4}{c}{Validation Seen} & \multicolumn{4}{c}{Validation Unseen} & \multicolumn{4}{c}{Test Unseen} \\
\cmidrule(lr){2-5}\cmidrule(lr){6-9}\cmidrule(lr){10-13}
Method & NE$\downarrow$ & SR$\uparrow$ & OSR$\uparrow$ & SPL$\uparrow$ & NE$\downarrow$ & SR$\uparrow$ & OSR$\uparrow$ & SPL$\uparrow$ & NE$\downarrow$ & SR$\uparrow$ & OSR$\uparrow$ & SPL$\uparrow$ \\
\midrule
Seq2Seq & 363.7 & 0.89 & 9.66 & 0.63 & 392.0 & 0.80 & 10.15 & 0.58 & 380.2 & 1.19 & 11.59 & 0.72 \\
Seq2Seq + \ours + LA & 224.7 & 6.42 & 14.40 & 3.80 & 235.8 & 5.93 & 13.93 & 3.60 & 223.3 & 6.98 & 16.47 & 3.65 \\
\midrule
CMA & 197.1 & 0.89 & 8.86 & 0.60 & 207.3 & 0.80 & 7.36 & 0.47 & 183.2 & 1.04 & 8.13 & 0.49 \\
CMA + \ours + LA & 168.6 & 8.91 & 16.39 & 4.45 & 182.8 & 6.88 & 13.17 & 3.25 & 157.1 & 9.19 & 16.08 & 3.85 \\
\midrule
MGP & 54.7 & 9.61 & 35.86 & 8.82 & 64.8 & 6.36 & 24.22 & 5.91 & 60.5 & 8.46 & 31.21 & 8.01 \\
MGP + \ours + LA & 31.3 & 50.52 & 59.78 & 41.90 & 43.3 & 39.38 & 47.62 & 30.98 & 35.8 & 48.31 & 56.85 & 38.87 \\
\midrule
FlightGPT & 65.0 & 16.90 & 31.85 & 14.98 & 64.7 & 16.18 & 31.21 & 14.65 & 60.2 & 24.24 & 41.55 & 21.95 \\
FlightGPT + \ours + LA & 49.2 & 44.05 & 54.53 & 34.16 & 48.7 & 42.40 & 54.85 & 32.44 & 46.9 & 48.85 & 61.24 & 38.18 \\
\midrule
HETT & 38.3 & 29.66 & 47.59 & 25.00 & 53.5 & 17.71 & 34.35 & 14.60 & 42.5 & 25.94 & 47.42 & 21.17 \\
HETT + \ours + LA & 27.6 & 57.59 & 69.27 & 44.73 & 38.7 & 46.25 & 57.36 & 33.93 & 30.1 & 54.55 & 67.49 & 39.96 \\
\midrule
LM-centroid & 43.6 & 25.52 & 32.16 & 24.71 & 43.2 & 23.79 & 30.55 & 23.10 & 31.8 & 40.20 & 46.26 & 38.77 \\
LM-centroid + \ours + LA & 30.6 & 53.53 & 61.34 & 46.33 & 31.2 & 50.61 & 60.27 & 42.72 & 24.3 & 61.37 & 72.11 & 53.49 \\
\bottomrule
\end{tabular}
\caption{\textbf{Results on all CityNav splits} in the format of prior work (revised annotations). Validation: held-out episodes. Seq2Seq and CMA: episodes kept by the official VLN-CE loader (2,257 / 2,513 / 5,117).}
\label{tab:full}
\end{table*}

\begin{table*}[t]
\centering
\footnotesize
\setlength{\tabcolsep}{3.5pt}
\renewcommand{\arraystretch}{1.1}
\begin{tabular}{lrrrrrrrrrrrr}
\toprule
& \multicolumn{4}{c}{Easy} & \multicolumn{4}{c}{Medium} & \multicolumn{4}{c}{Hard} \\
\cmidrule(lr){2-5}\cmidrule(lr){6-9}\cmidrule(lr){10-13}
Method & NE$\downarrow$ & SR$\uparrow$ & OSR$\uparrow$ & SPL$\uparrow$ & NE$\downarrow$ & SR$\uparrow$ & OSR$\uparrow$ & SPL$\uparrow$ & NE$\downarrow$ & SR$\uparrow$ & OSR$\uparrow$ & SPL$\uparrow$ \\
\midrule
Seq2Seq & 430.3 & 2.30 & 17.21 & 1.24 & 379.0 & 0.49 & 7.71 & 0.39 & 311.4 & 0.41 & 8.04 & 0.35 \\
Seq2Seq + \ours + LA & 261.0 & 9.41 & 23.19 & 3.93 & 218.8 & 4.81 & 11.72 & 2.81 & 175.5 & 5.98 & 12.36 & 4.17 \\
\midrule
CMA & 165.2 & 1.37 & 14.31 & 0.25 & 177.1 & 0.99 & 6.11 & 0.76 & 215.1 & 0.62 & 1.72 & 0.52 \\
CMA + \ours + LA & 134.0 & 10.98 & 24.22 & 2.14 & 152.0 & 7.83 & 12.28 & 4.44 & 195.2 & 8.17 & 8.93 & 5.58 \\
\midrule
MGP & 62.4 & 7.26 & 48.68 & 6.29 & 60.2 & 9.12 & 25.74 & 8.94 & 58.3 & 9.40 & 13.34 & 9.33 \\
MGP + \ours + LA & 33.7 & 52.67 & 68.24 & 37.04 & 33.1 & 50.89 & 56.85 & 44.97 & 41.5 & 39.49 & 41.26 & 34.65 \\
\midrule
FlightGPT & 49.0 & 26.67 & 48.53 & 23.12 & 60.1 & 23.60 & 39.69 & 21.80 & 75.7 & 21.62 & 34.03 & 20.53 \\
FlightGPT + \ours + LA & 36.0 & 51.61 & 67.28 & 35.78 & 46.3 & 48.33 & 59.06 & 39.63 & 62.5 & 45.66 & 55.39 & 39.86 \\
\midrule
HETT & 40.4 & 28.83 & 52.28 & 20.69 & 43.6 & 24.73 & 45.71 & 21.52 & 44.1 & 23.32 & 42.64 & 21.43 \\
HETT + \ours + LA & 29.4 & 55.26 & 70.40 & 33.34 & 31.2 & 52.56 & 64.60 & 41.22 & 29.8 & 55.78 & 66.69 & 47.63 \\
\midrule
LM-centroid & 30.6 & 41.23 & 50.02 & 38.98 & 33.6 & 37.78 & 42.67 & 36.79 & 31.3 & 41.46 & 45.07 & 40.68 \\
LM-centroid + \ours + LA & 23.5 & 60.12 & 73.67 & 47.90 & 25.6 & 59.95 & 68.71 & 54.00 & 23.9 & 64.65 & 73.72 & 60.56 \\
\bottomrule
\end{tabular}
\caption{\textbf{Results on the three difficulty levels of test-unseen} defined by CityNav (2,081 / 1,678 / 1,522 episodes with the revised annotations).}
\label{tab:difficulty}
\end{table*}

\end{document}